\documentclass[letterpaper]{article} 
\usepackage{aaai23}  
\usepackage{times}  
\usepackage{helvet}  
\usepackage{courier}  
\usepackage[hyphens]{url}  
\usepackage{graphicx} 
\usepackage{natbib}  
\usepackage{caption} 
\usepackage{algorithm}
\usepackage{algorithmic}
\usepackage{mathrsfs}
\usepackage{amsmath}
\usepackage{newfloat}
\usepackage{dblfloatfix}
\usepackage{mathrsfs}
\usepackage{listings}
\usepackage{mathrsfs}
\usepackage{xcolor}
\definecolor{bluetime}{RGB}{25,93,171}
\definecolor{purunet}{RGB}{157, 50, 163}
\DeclareCaptionStyle{ruled}{labelfont=normalfont,labelsep=colon,strut=off} 
\floatstyle{ruled}
\newfloat{listing}{tb}{lst}{}
\floatname{listing}{Listing}
\usepackage{fancyhdr}
\usepackage{subcaption}
\fancypagestyle{firstpagehf}
{
   \fancyhf{}
   \fancyhead[HC]{The Second Workshop on Deployable AI $@$ AAAI 2024}
   \fancyfoot[C]{\thepage}
}

\title{Unraveling the Temporal Dynamics of the Unet in Diffusion Models}
\author{
    Vidya Prasad$^1$, Chen Zhu-Tian$^2$, Anna Vilanova$^1$, Hanspeter Pfister$^2$ \\Nicola Pezzotti$^{1,3}$, Hendrik Strobelt$^{4}$
}
\affiliations {
    \textsuperscript{\rm 1} Eindhoven University of Technology,
    \textsuperscript{\rm 2} Harvard University,
    \textsuperscript{\rm 3} Philips Medical Systems,
    \textsuperscript{\rm 4} IBM Research\quad\\
}

\usepackage{bibentry}
\usepackage{eucal}

\begin{document}
\thispagestyle{firstpagehf}
\maketitle

\begin{abstract}
Diffusion models have garnered significant attention since they can effectively learn complex multivariate Gaussian distributions, resulting in diverse, high-quality outcomes. They introduce Gaussian noise into training data and reconstruct the original data iteratively.
Central to this iterative process is a single Unet,  adapting across time steps to facilitate generation. Recent work revealed the presence of composition and denoising phases in this generation process, raising questions about the Unets' varying roles. Our study dives into the dynamic behavior of Unets within denoising diffusion probabilistic models (DDPM), focusing on (de)convolutional blocks and skip connections across time steps. 
We propose an analytical method to systematically assess the impact of time steps and core Unet components on the final output. This method eliminates components to study causal relations and investigate their influence on output changes.
The main purpose is to understand the temporal dynamics and identify potential shortcuts during inference. Our findings provide valuable insights into the various generation phases during inference and shed light on the Unets' usage patterns across these phases. Leveraging these insights, we identify redundancies in GLIDE (an improved DDPM) and improve inference time by $\sim$$27\%$ with minimal degradation in output quality. 
Our ultimate goal is to guide more informed optimization strategies for inference and influence new model designs. 
\end{abstract}

\section{Introduction}

Diffusion models learn complex multivariate Gaussian distributions, enabling high-quality and diverse outcomes. They have found several applications including generation~\cite{nichol2021glide}, inverse problems~\cite{inverseprob}, segmentation~\cite{segmentation}, and anomaly detection~\cite{wyatt2022anoddpm}. During training, these models introduce controlled corruptions, like Gaussian noise, followed by an iterative de-corruption process to reconstruct the original data. This reconstruction process employs models like the Unet~\cite{ronneberger2015u} per iteration, enabling effective learning. This Unet takes each intermediate noisy image and the time step as input and produces an intermediate output to generate data from noise.

The inference time complexity hinders diffusion models from being used in time-critical applications in the real world~\cite{yang2022diffusionsurvey}. Researchers have explored ways to enhance efficiency while maintaining quality, including faster sampling methods~\cite{ddim, nichol2021improved, kong2021fast, karras2022elucidating} to reduce the time steps. 
Architectures to process time steps in parallel~\cite{zheng2023fast, shih2023parallel} require more compute power or tuning, which is challenging to implement in large, well-established models~\cite{nichol2021glide, stablediff}. 

Prior work~\cite{deja2022analyzing} exploring the generation process in diffusion models have highlighted distinct composition and denoising phases, showing that time steps contribute variably to the final output. This sparks fundamental questions about the varying purposes of time steps and the Unets' adaptable role.
Unets, with their skip connections and encoder-decoder structure, have inherent dynamism and can employ different connections across inputs, suitable for various applications. Yet, the precise utility of this adaptable Unet and its components, i.e., the (de)convolutional blocks and skip connections across time steps, remains unknown. 
Understanding this adaptability over time can offer insights into the transitions between different generative phases, thereby supporting model inference and design optimizations.
Hence, we delve deeper into the diffusion inference process to deepen our understanding and guide the development of informed optimization strategies.

We present an analytical method to evaluate each time step's impact and the core Unet's utility on the final output of denoising diffusion probabilistic models (DDPMs).
We systematically eliminate time steps and Unet components and analyze their effects on the output to elucidate their purpose in the generation process. Our findings enhance the understanding of the image formation process and Unet's role. By identifying areas of redundancy, we aim to identify shortcuts in pretrained diffusion models crucial for time-sensitive applications. Our contributions are listed below.

\begin{itemize}
    \item We introduce an analytical method to assess the impact of time steps and the Unet in diffusion models.
    \item We investigate the causal relationship between Unet interventions and time steps on model outputs, elucidating the image formation process and Unet's role.
    \item By identifying areas of redundancy, we uncover shortcut possibilities during inference.
\end{itemize}

\section{Background}
Denoising diffusion probabilistic models~\cite{sohl2015deep, ho2020denoising, nichol2021improved} aim to approximate the true distribution $q(\textbf{x}_0)$ of training data. To do so, each sample from $q(\textbf{x}_0)$ is gradually degraded by iteratively adding Gaussian noise, resulting in a set of latent variables, $\textbf{x}_1, \textbf{x}_2, ..., \textbf{x}_T$, given by, 

\begin{equation}
\label{eq:noiseadd}
q(\textbf{x}_t|\textbf{x}_{t-1}):= (\textbf{x}_t; \sqrt{\alpha_t}\textbf{x}_{t-1}, (1-\alpha_t)I)
\end{equation}
\newline
This sequential introduction of noise, from $\textbf{x}_0$ to $\textbf{x}_t$, is called the forward process. 
A model $p_\theta(\textbf{x}_{t-1}|\textbf{x}_{t})$ can be trained to approximate the true posterior distribution $p_\theta(\textbf{x}_{t-1}|\textbf{x}_{t}) := \mathcal{N}(\mu_\theta(\textbf{x}_t), \sigma_\theta(\textbf{x}_t))$. This trained model generates samples starting from Gaussian noise, i.e., $\textbf{x}_T$$\sim$$\mathcal{N}(0, I)$ and progressively reduces this noise iteratively through $\textbf{x}_{T-1}, ..., \textbf{x}_0$. In practice, $\textbf{x}_t$ sampled from $q(\textbf{x}_t|\textbf{x}_0)$ are generated by applying Gaussian noise $\epsilon$ to $\textbf{x}_0$. A model $\epsilon_\theta$ is then trained to predict the noise introduced during the forward process using mean-squared error loss, defined as,

\begin{equation}
\label{eq:mse}
    L:= E_{t\sim[1,T], \textbf{x}_0\sim q(x_0), \epsilon\sim\mathscr{N}(0,I)}[\lVert \epsilon - \epsilon_\theta(\textbf{x}_t, t) \rVert^2 ]\\
\end{equation}
\newline
The trainable model $\epsilon_\theta$ that predicts the noise at each time step $t$, is commonly implemented as a Unet. The noise $\epsilon_\theta(\textbf{x}_t, t)$ predicted by the model is utilized to generate $\textbf{x}_{t-1}$ from $\textbf{x}_t$ as follows,

\begin{equation}
\begin{aligned}
\label{eq:xtpred}
\textbf{x}_{t-1} = \sqrt{\textcolor{bluetime}{\alpha_{t-1}}}(\dfrac{\textbf{x}_t - \sqrt{1-\alpha_t}\textcolor{purunet}{\epsilon_\theta}(\textbf{x}_t, t)}{\sqrt{\alpha_t}}) 
\\+ \sqrt{1-\textcolor{bluetime}{\alpha_{t-1}}-\sigma^2_t} . \textcolor{purunet}{\epsilon_\theta}(\textbf{x}_t, t) 
\\+ \sigma_t \epsilon_t
\end{aligned}
\end{equation}

Where $\sigma_t$ is a constant or dependent on $\sigma_\theta(\textbf{x}_t)$. The first term is the predicted $\textbf{x}_0$ at time step $t$, given by $\textbf{x}^t_{est_0}$. The second term is the direction pointing back to $\textbf{x}_t$, and the last is the random noise added. The colors in Equation~\ref{eq:xtpred} are the parts we modify to intervene on \textcolor{bluetime}{time steps} and \textcolor{purunet}{Unet connections}. We detail our intervention method in later sections. 

\section{Related work}

The large inference time in diffusion models has prompted advancements in new architectures and efficient sampling methods to enhance speed without compromising quality. 

Among these, denoising diffusion implicit models or DDIMs~\cite{ddim, zhang2023gddim} are a more efficient non-Markovian sampling strategy than Markovian sampling in DDPMs~\cite{ho2020denoising}. DDIMs offer significantly accelerated ($10$-$50$x) high-quality sample generation. However, DDIMs demonstrate inferior sample outcomes when used directly with the ancestral sampling. They only achieve better quality when corrector steps, necessitating manual parameter tuning.
Improved DDPMs or iDDPMs~\cite{nichol2021improved} focus on learning variances during reconstruction, enabling sample generation with significantly fewer time steps and minimal degradation in quality. iDDPMs support rapid sampling directly from the ancestral sampling process, removing the need for fine-tuning additional parameters.
Further, explorations into solvers with adaptive step sizes tailored for score-based diffusion models~\cite{jolicoeur-martineau2022gotta} have surfaced for a piece-by-piece optimization. 
Additionally, higher-order ordinary differential equations (ODE) solvers have been explored~\cite{karras2022elucidating} and have provided valuable insights into optimizing inference times. 
Building upon these works, diffusion exponential integrator sampler or DEIS~\cite{deis} leverages specific discretization methods from ODEs for further speed and sample quality improvements. 

Most of these fast sampling methods are sequential in nature and try to reduce the number of time steps for time efficiency. Researchers have hence explored parallelizing time steps. 
Sampling latency was improved via Parallel Diffusion Generative Model Sampling, ParaDiGMS~\cite{shih2023parallel}. It aims to guess the full denoising or generation path and iteratively refine this until convergence. These refinement steps were much smaller than the time steps in diffusion models, allowing for faster sampling with much higher computational costs. Similarly, diffusion model sampling
with neural operator (DSNO) also aims to model the denoising trajectory via temporal
convolution layers added into the core diffusion model. While DSNO claims to have small additions to complexity, the method involves training that can be complex to implement in well-established models~\cite{nichol2021glide, stablediff}.

\begin{figure*}[!b]
\renewcommand{\thefigure}{2}
    \centering
    \includegraphics[width=0.85\textwidth]{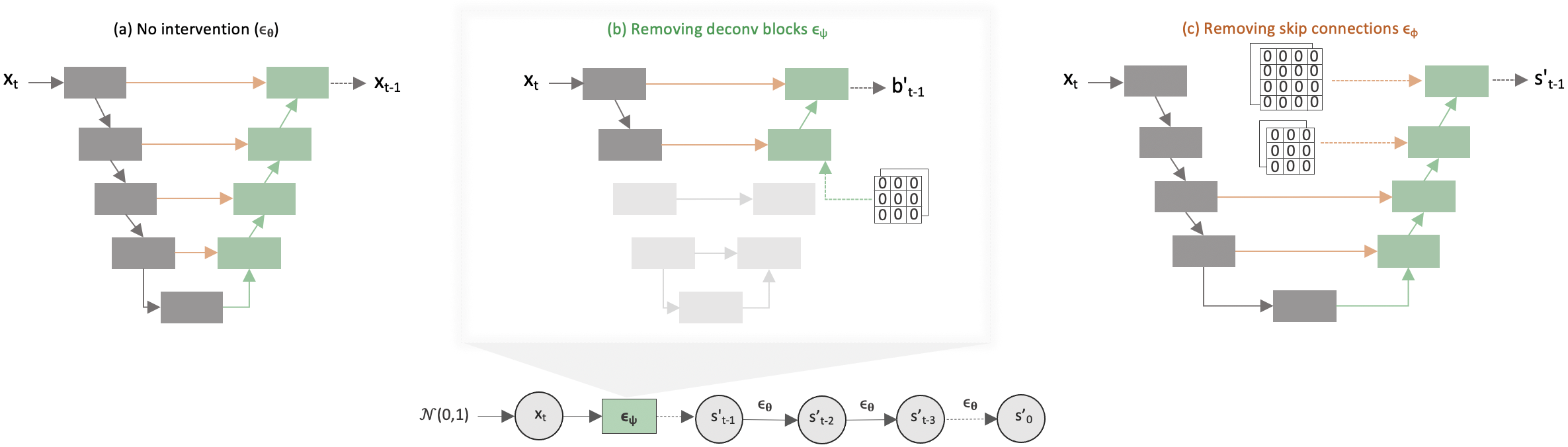}
    \caption{(a) The Unet model $\epsilon_\theta$ in a regular setting. (b) $nb=2$ deconvolutional blocks removed in a single time step ($n=1$) as shown in the bottom row generation sequence. (c) $ns=2$ skip connections removed in a single time step. The missing activations are replaced with a map of zeros of the same shape as the connection removed.}
    \label{fig:unet_intervene_method}
\end{figure*}
While the parallel architectures lead to an increase in computation and require some training, the sequential fast sampling methods treat all time steps similarly. 
Recent works~\cite{deja2022analyzing} delved into the iterative generative process within diffusion models, highlighted the presence of composition and denoising phases, and used these insights for a two-part continual learning. 
Prompted by this understanding of the distinct phases, we question whether all time steps are equal and warrant the same treatment. 
This study aims to build upon these insights to unravel the temporal dynamics within diffusion models. Specifically, we aim to examine the roles played by time steps and Unet connections in the generation process, which can pave the way for more informed model optimization and possibly even design.

\section{Method}
\begin{figure}
\renewcommand{\thefigure}{1}
    \centering
    \includegraphics[width=0.99\linewidth]{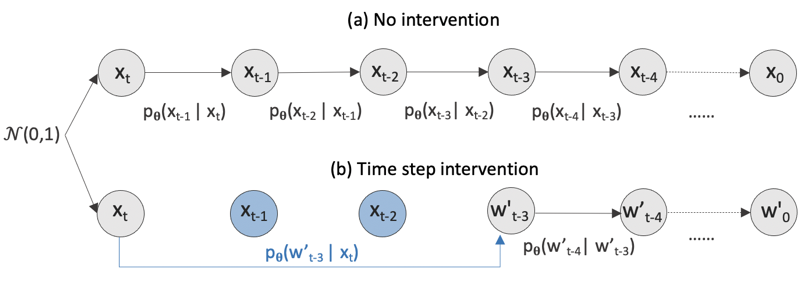}
    \caption{(a) The inference process in a regular setting. (b) Interventions on specific time steps ($t-1$, $t-2$) are shown in blue. The corresponding intervened ($\textbf{w}'_0$), and original ($\textbf{x}_0$) are compared to identify causal relations.}
    \label{fig:time_intervene_method}
\end{figure}

The learned diffusion model $\epsilon_\theta(\textbf{x}_t, t)$ is iteratively applied from $t=T$ to $t=0$ to generate the final output $\textbf{x}_0$. The main idea of our analytical method is to make an intervention or manipulation within a time step. This intervention is introduced in one or more time steps $t\in{0,...T}$. By scrutinizing the differences between the original $\textbf{x}_0$ and intervened output, for example, $\textbf{w}'_0$ in Figure~\ref{fig:time_intervene_method}, we aim to establish the causal relations between the intervention and their utility and impact on the output. The interventions include various manipulations, including removing entire time steps, (de)convolutional blocks, and skip connections within the Unet. Each intervention provides insight into the functional significance of the component.
\subsubsection{Interventions on time steps:} To identify the significance of each time step and determine the transition between the composition and denoising parts, we intervene on entire time steps (see Figure~\ref{fig:time_intervene_method}).
Time steps are removed starting from $t_{start}=t$ in the generation process and continue until $t-n$. $\textbf{w}'_{t-n}$ is the corresponding intervened output. Removing time steps essentially implies that rather than estimating $\textbf{x}_{t-1}$ via the regular process (see Figure~\ref{fig:time_intervene_method}a), we estimate $\textbf{x}_{t-n}$ from time step $t$ directly, i.e., $\textbf{w}'_{t-3}$ in Figure~\ref{fig:time_intervene_method}b. In this case, two time steps are removed. This can be generalized to $n$ time steps. We compute $\textbf{w}'_{t-n}$ from $\textbf{x}_{t}$ as follows,

\begin{equation}
\begin{aligned}
\textbf{w}'_{t-n} = \sqrt{\textcolor{bluetime}{\alpha_{t-n}}}(\dfrac{\textbf{x}_t - \sqrt{1-\alpha_t}\epsilon_\theta(\textbf{x}_t, t)}{\sqrt{\alpha_t}}) 
\\+ \sqrt{1-\textcolor{bluetime}{\alpha_{t-n}}-\sigma^2_t} . \epsilon_\theta(\textbf{x}_t, t) 
\\+ \sigma_t \epsilon_t
\end{aligned}
\end{equation}
Once the $\textbf{w}'_{t-n}$ is computed, $\textbf{w}'_{t-n-1}$...$\textbf{w}'_0$ follows the regular Equation~\ref{eq:xtpred}. In the example in Figure~\ref{fig:time_intervene_method}, the intervention starts after $t_{start}=T$ until $T-3$, i.e., $\textbf{w}'_{T-3}$.
\subsubsection{Interventions on Unet components:}
Within a time step $t$, we delve into the utility of the Unet components. Since layers higher up in the ``U'' contribute more to local features due to their smaller receptive field, and vice-versa, we hypothesize that not all connections of the Unet hold equal significance at every time step. 
Hence, we expect a trend in their utility concerning the phase of the generation process in diffusion models.

We perform interventions on the Unet ($\epsilon_{\theta}$) for a specified range of time steps. Specifically, we intervene on components, including the skip connections and the Unet deconvolutional blocks. 
This skip connection intervened Unet ($\epsilon_{\phi}$) is specified by the number of skip connections ($ns$) removed from the top of the Unets starting at time step $t_{start}$ until $t_{start}-n$. The weights of $\epsilon_{\phi}$ are the same as $\epsilon_{\theta}$. The skip intervention resets the activations of the top $ns$ skip connections to zeros, as shown in Figure~\ref{fig:unet_intervene_method}c. This activation reset draws inspiration from existing intervention methods~\cite{bau2018gan}. 
The standard Equation~\ref{eq:xtpred} to generate $\textbf{x}_{t-1}$ based on predicted noise at $t$ is modified; the intervened Unet $\textcolor{purunet}{\epsilon_{\phi}}$ is used instead of $\textcolor{purunet}{\epsilon_{\theta}}$ to predict the noise (see~Equation~\ref{eq:spred}).
The intervened stage output and subsequent outputs are denoted by $\textbf{s}'_{t-1},...\textbf{s}'_0$. 
\begin{equation}
\begin{aligned}
\label{eq:spred}
\textbf{s}'_{t-1} = \sqrt{\alpha_{t-1}}(\dfrac{\textbf{x}_t - \sqrt{1-\alpha_t}\textcolor{purunet}{\epsilon_\phi}(\textbf{x}_t, t)}{\sqrt{\alpha_t}}) 
\\+ \sqrt{1-\alpha_{t-1}-\sigma^2_t} . \textcolor{purunet}{\epsilon_\phi}(\textbf{x}_t, t) 
\\+ \sigma_t \epsilon_t
\end{aligned}
\end{equation}

Similar to the skip intervention, we define interventions on deconvolution blocks in the Unet ($\epsilon_{\psi}$), specified by the number of deconvolutional blocks ($nb$) removed from the bottom of the Unets in the time step $t_{start}$ until $t_{start}-n$.
We set the output activations of the $nb^{th}$ block from the bottom of the ``U'' to zeros (see Figure~\ref{fig:unet_intervene_method}b). This implies that the bottom $nb$ blocks on the encoder and the decoder side do not need to be processed. 
The intervened stage output and subsequent outputs are denoted by $\textbf{b}'_{t-1},...\textbf{b}'_0$. Interventions can be applied on $n$ timesteps, for example, $n=1$ in the bottom row of Figure~\ref{fig:unet_intervene_method}b.
The original $\textbf{x}_0$ is compared with the intervened $\textbf{s}'_0$ or $\textbf{b}'_0$ to explore causal relations between the Unet components and the output.
We compute the peak signal-to-noise ratio (PSNR) and structural similarity index (SSIM) between intervened outputs ($\textbf{w}'_0$, $\textbf{b}'_0$, or $\textbf{s}'_0$) and non-intervened outputs ($\textbf{x}_0$) to discern trends. Note that the same noise seed is used for a specific instance of the intervened ($\textbf{w}'_0$, $\textbf{b}'_0$, or $\textbf{s}'_0$) and original $\textbf{x}_0$ to enable comparison.

\section{Experiments \& Findings}

For our experiments, we use the same set of $100$ samples generated from $100$ prompts, one sample per prompt. Randomly selected ImageNet classes are used as prompts. We use the pre-trained GLIDE~\cite{nichol2021glide} text-to-image generator. GLIDE has a base improved DDPM (iDDPM) model to generate samples of size 64x64 in $100$ spaced time steps. These generation steps are $t\in{100,..,2,1}$. GLIDE has a Unet with $16$ convolutional blocks. We show how even this simpler, faster model has redundancies.
\begin{figure}[!t]
    \centering
    \includegraphics[width=0.95\linewidth]{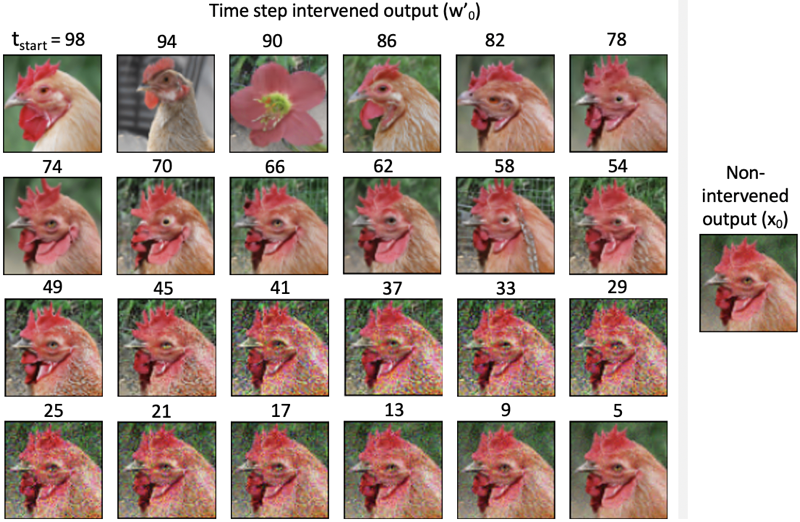}
    \caption{The time step intervened output $\textbf{w}'_0$ when $n=5$ consecutive steps are eliminated at different $t_{start}$ values.  The top row shows significant semantic alterations; the second row leads to discernible yet smaller semantic changes; the third row shows coarse denoising with semantic variation; and the last row indicates pure denoising effects.   }
    \label{fig:timestep5_del}
\end{figure}

\subsection{Exploration of generation phases}
Here, we focus on the significance of time steps in diffusion models.
Initially, we explore removing individual time steps ($n=1$) at varying $t_{start}$ values. 
This systematic elimination of each time step yields a collection of T outputs. We observe that modifying a single time step does not substantially alter the model's output, demonstrating its self-correction capabilities and resilience to minor perturbations.

To create more perceivable changes, we eliminate $n=5$ consecutive time steps starting at different $t_{start}$ values. Interestingly, even this small modification has a pronounced influence on the model's output $\textbf{w}'_0$ (see Figure~\ref{fig:timestep5_del}). 
Our exploration reveals distinctive patterns based on eliminating specific time steps.
Eliminating early time steps, representing initial generation stages, resulted in significant semantic alterations in the output (see the top row in Figure~\ref{fig:timestep5_del}), coinciding with the beta schedule employed where the noise at these steps is significantly higher than the rest. 
Subsequent time steps exhibited a shift towards coarse noise and minor semantic effects (see the middle rows of Figure~\ref{fig:timestep5_del}), while the later stages demonstrated substantial denoising efforts (see the bottom row of Figure~\ref{fig:timestep5_del}).  

\begin{figure}[!ht]
    \centering
    \includegraphics[width=0.95\linewidth]{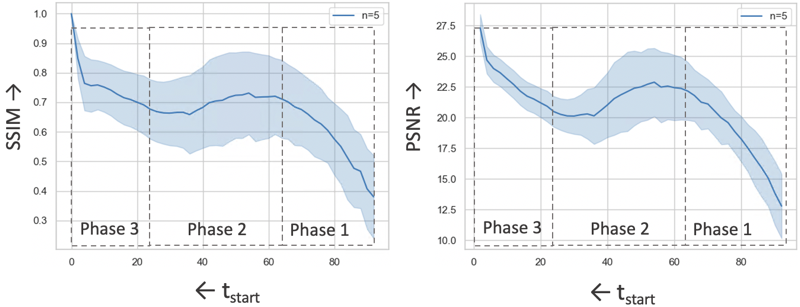}
    \caption{SSIM (left) and PSNR (right) between the original $\textbf{x}_0$ and the time step intervened $\textbf{w}'_0$ observed after removing $n=5$ time steps starting at $t_{start}$ until $t_{start}-5$. The mean $\pm$ 1 standard deviation across 100 instances is shown.}
    \label{fig:stages_timestep}
\end{figure}

 Figure~\ref{fig:stages_timestep} shows the SSIM and PSNR trends between the intervened $\textbf{w}'_0$ and original $\textbf{x}_0$ across different $t_{start}$ values. It highlights three discernible patterns. Phase 1 demonstrates significant semantic changes due to poor SSIM, underscoring a primary composition phase. 
Subsequently, in phase 2, the model transitioned from a pure generative to a denoising phase with some semantic changes. This transition is marked by decreased PSNR and SSIM as the generated images exhibited more noise. Finally, the last phase shows a notable increase in PSNR and SSIM, signifying a predominantly denoising phase in the image generation process.

\subsection{Utility of Unet connections}
\begin{figure}[!b]
    \centering
    \includegraphics[width=0.95\linewidth]{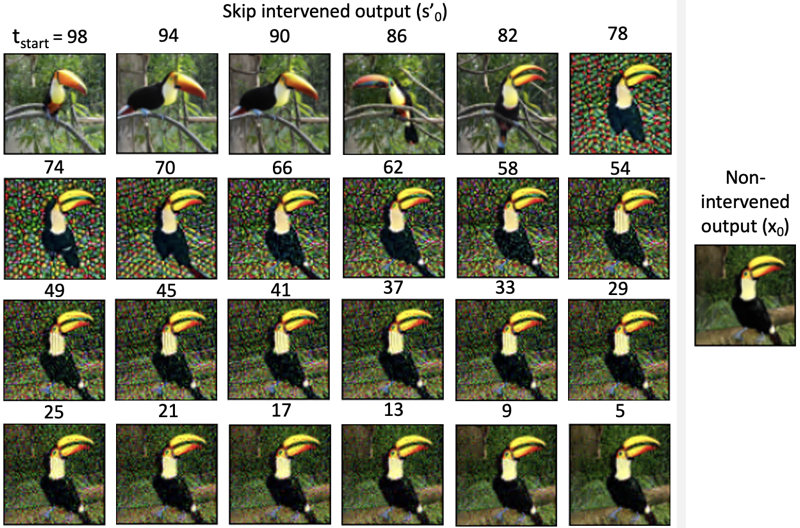}
    \caption{The resulting output $\textbf{s}'_0$ after removing the top two skip connections ($ns=2$) for five consecutive time steps ($n=5$) at various $t_{start}$ values. The top row indicates evident semantic changes upon removing this connection in the initial generation steps. Subsequent steps reveal its contribution to coarse and fine denoising effects.}
    \label{fig:top_1_single}
\end{figure} 
Given that the same Unet is used through the phases in Figure~\ref{fig:stages_timestep}, we hypothesize a temporal variance in the utility of the Unet's connections.
We expect that not all connections of the Unet are important in all stages.  For example, in the later stages (Phase 3), the reliance on lower bottleneck features might drop significantly.
These deeper layers, having larger receptive fields, likely prioritize semantic changes, potentially leading to a shift in their significance during Phase 3, where the focus shifts towards denoising. This hypothesis prompts us to examine whether the Unet contains redundant connections, especially in phases 2 and 3. 
\begin{figure}[!t]
    \centering
    \includegraphics[width=0.95\linewidth]{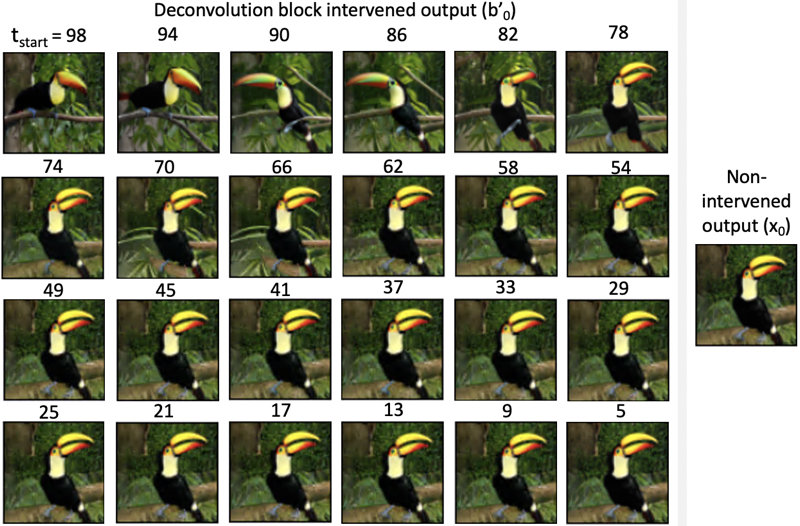}
    \caption{The intervened output $\textbf{b}'_0$ after removing the bottom two deconvolutional blocks ($nb=2$) for $n=5$ consecutive time steps at various $t_{start}$ values. The top row indicates evident semantic changes upon removing this connection in the initial generation steps. Subsequent steps reveal minimal contributions to the output. }
    \label{fig:bot_1_single}
\end{figure}
\begin{figure}[!b]
    \centering
    \includegraphics[width=0.95\linewidth]{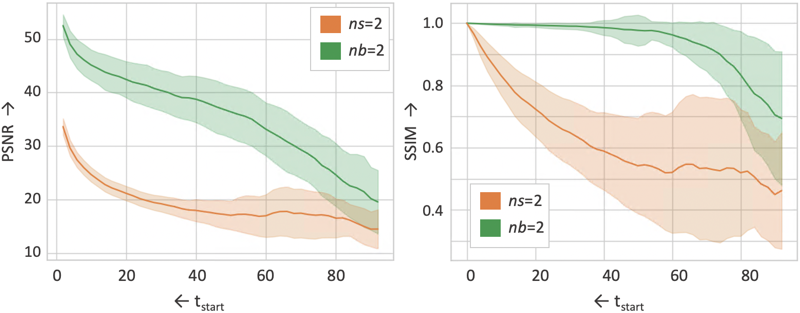}
    \caption{PSNR (left) and SSIM (right) between the original $\textbf{x}_0$ and the deconvolution block intervened $\textbf{b}'_0$ vs. the skip intervened $\textbf{s}'_0$ output observed after removing $n=5$ time steps starting at a step $t_{start}$ until $t_{start}-5$. The mean $\pm$ 1 standard deviation of metrics across 100 instances is shown. }
    \label{fig:deconvskip_intervene_ssimpsnr}
\end{figure}

\subsubsection{Purpose of connections:}
Exploring the difference in the purpose of the Unet connections, we remove the two top skip connections ($ns=2$) and two bottom deconvolutional blocks ($nb=2$). Analyzing the resultant intervened outputs ($\textbf{s}'_{0}$ vs. $\textbf{b}'_{0}$) across varied start time steps ($t_{start}$) reveals interesting insights. 
Initially, during Phase 1 of the generation process (top rows of Figures~\ref{fig:top_1_single} and~\ref{fig:bot_1_single}), both interventions induce semantic changes. However, as the generation progresses, the distinctions between the two become more apparent. The top skip connections seem to aid in denoising, while the lower connections primarily contribute to semantic alterations (seen in lower rows of Figures~\ref{fig:top_1_single} and~\ref{fig:bot_1_single}).
Interestingly, we also note in the lower rows of Figure~\ref{fig:bot_1_single} that these lower deconvolution block, primarily influencing semantic changes, becomes redundant after phase 1.

This trend is further supported by the PSNR and SSIM trends between $\textbf{s}'_{0}$ and $\textbf{x}'_{0}$ vs. $\textbf{b}'_{0}$ and $\textbf{x}'_{0}$ across different $t_{start}$ values. The PSNR values when removing skip connections (indicated by the orange line in Figure~\ref{fig:deconvskip_intervene_ssimpsnr}) are notably lower compared to removing deconvolution blocks (in green), reinforcing the skip connections' denoising contribution. Conversely, when exploring SSIM ($\textbf{s}'_{0}$ vs. $\textbf{x}'_{0}$ vs. $\textbf{b}'_{0}$ vs. $\textbf{x}'_{0}$), removing lower deconvolutional blocks shows minimal effects post Phase 1, possibly indicating their redundancy.

\begin{figure}
    \centering
    \includegraphics[width=0.99\linewidth]{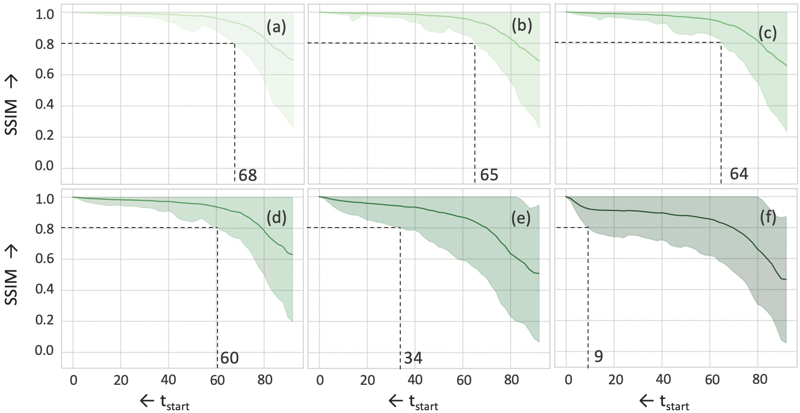}
    \caption{ SSIM between the original $\textbf{x}_0$ and the intervened deconvolutional block $\textbf{b}'_0$, observed after removing $nb$ blocks for $n=5$ time steps, starting from step $t_{start}$ to $t_{start}-5$. Shown are the mean $\pm$ $2$ standard deviations across 100 samples for (a) $nb=2$, (b) $nb=4$, (c) $nb=6$, (d) $nb=8$, (e) $nb=10$, and (f) $nb=12$. }
    \label{fig:unet_frombot}
\end{figure}

\subsubsection{Trends in utility:}
Since the lowest deconvolutional block becomes redundant toward the end of phase 1, we investigate whether these blocks from the bottom up gradually diminish in importance over the generation time steps. We remove varying numbers of deconvolutional blocks ($nb\in{2,4,6,8,10,12}$) from the bottom and compare the intervened ($\textbf{b}'_0$) and non-intervened ($\textbf{x}_0$) outputs (see Figure~\ref{fig:unet_frombot}). 
We use an SSIM threshold of 0.8 to mark visual differences between $\textbf{b}'_0$ and $\textbf{x}_0$, found empirically through visual exploration and analysis. This threshold efficiently highlighted differences in global semantics, effectively supporting our pursuit of insights. While this threshold serves our purpose, it is adaptable to the needs of specific applications.

Figures~\ref{fig:unet_frombot}a-\ref{fig:unet_frombot}f depict a gradual improvement in SSIM (right to left) in the generation process for different $nb$ values. 
Removing fewer blocks achieves perceptual similarity with non-intervened outputs sooner than higher block counts. 
For example,  $nb=2$ achieves an SSIM of $0.8$ sooner than $nb=4$ (at $t_{start}\approx68$), $nb=4$ reaches the same threshold earlier than $nb=6$ (at $t_{start}\approx65$), and so forth. 
This pattern suggests a gradual reduction in the significance of these blocks bottom-up over time.

\subsection{Limitation in consecutive intervention time steps}
In previous experiments, we explored interventions across $n=5$ consecutive time steps.
For example, removing the bottom two deconvolution blocks ($nb=2$) starting at time step $t_{start}\approx70$ for $n=5$ steps had minimal impact on the final image $\textbf{b}'_0$ with $ssim (\textbf{b}'_0, \textbf{x}_0)\ge0.8$.
\begin{figure}[!t]
    \centering
    \includegraphics[width=0.9\linewidth]{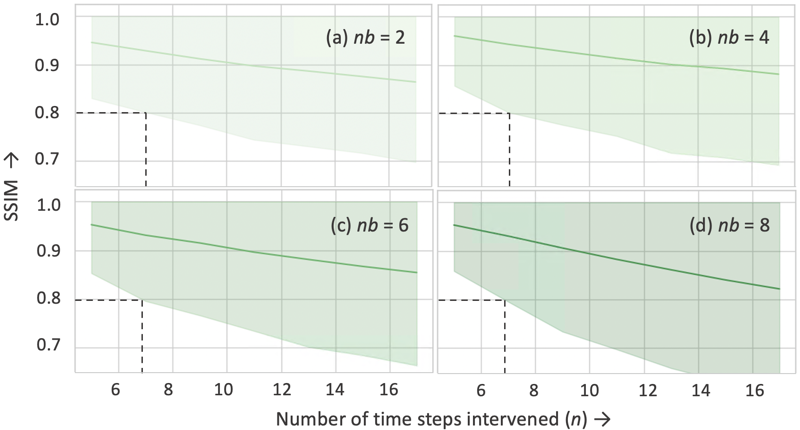}
    \caption{SSIM between the original $\textbf{x}_0$ and the intervened $\textbf{b}'_0$ after the removal of $nb$ blocks across $n$ steps. Shown are the mean $\pm$ $2$ standard deviations over 100 instances for (a) $nb=2$ at $t_{start}=70$, (b) $nb=4$ at $t_{start}=65$, (c) $nb=6$ at $t_{start}=65$, and (d) $nb=8$ at $t_{start}=60$.}
    \label{fig:window_exp}
\end{figure}

Expanding further, we aim to assess the boundaries concerning the number of time steps $n$ that can be intervened on before observing an effect on the final output, i.e., at what $n$ is the $ssim (\textbf{b}'_0, \textbf{x}_0)<0.8$. Intervened outputs $\textbf{b}'_0$ are computed for varying $n$ time steps and $nb$ blocks, as depicted in Figure~\ref{fig:window_exp}. The starting intervention step $t_{start}$ for each of the $nb$ blocks removed was determined based on the point at which each block becomes redundant from Figure~\ref{fig:unet_frombot}.
Figure~\ref{fig:window_exp} indicates a limitation; no more than $n=7$ time steps can be removed without compromising the output. This finding suggests that the model possesses a self-correction mechanism up to a certain threshold, beyond which its performance is noticeably compromised.

\subsection{Cut-relax-cut strategy}
\begin{figure}
    \centering
    \includegraphics[width=0.95\linewidth]{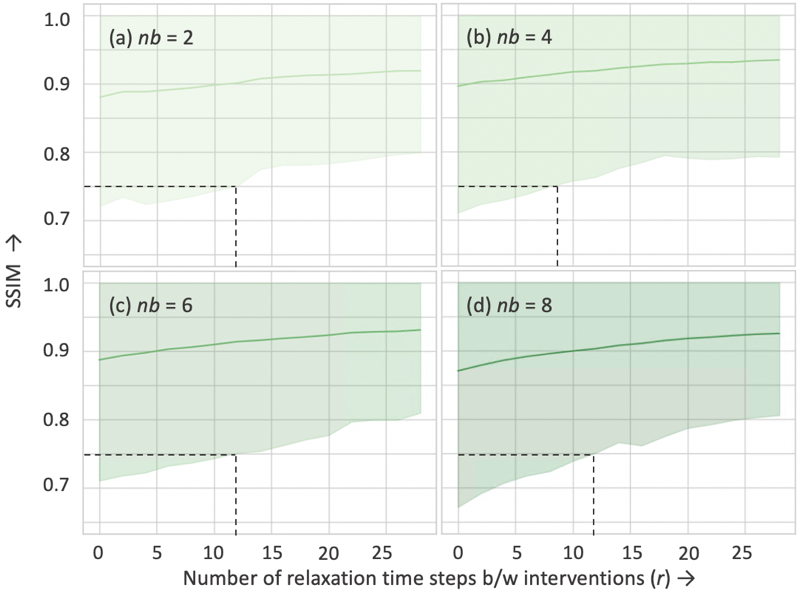}
    \caption{SSIM between the original $\textbf{x}_0$ and the deconvolution block intervened $\textbf{b}'_0$ observed after removing $nb$ blocks for $n=7$ time steps, a relaxation period of $r$ steps, followed by a second removal of $nb$ blocks for $n=7$ time steps. Shown are the mean $\pm$ $2$ standard deviations over 100 instances for (a) $nb=2$ and $t_{start}=70$ (b) $nb=4$ and $t_{start}=65$, (c) $nb=6$ and $t_{start}=65$, (d) $nb=8$ and  $t_{start}=60$. }
    \label{fig:cut-relax-cut}
\end{figure}
Given the limitation on the consecutive intervention steps $n$, we explore means to further optimize the model. 
Based on the redundancy patterns and self-correction capabilities of the model to deviate back to the mean noise, we explore a ``cut-relax-cut" strategy. This involves first cutting connections for $n$ steps until the intervened output starts to have an impact. We then allow the model to ``relax" for $r$ time steps without intervention to enable recovery. Following this, we assess the model's response to further interventions.

We continue our exploration based on our findings concerning the start time step $t_{start}$ and the number of intervention steps $n$ across various instances of $nb$ blocks removed, as depicted Figures~\ref{fig:unet_frombot} and~\ref{fig:window_exp}.
Since there are limitations in the number of time steps $n$ that can be removed, we investigate whether allowing the model to undergo a relaxation period of $r$ steps might enable subsequent intervention for an additional $n$ steps.
In Figure~\ref{fig:cut-relax-cut}, we observe that with $r \in [15,20]$, we can indeed intervene for another $n$ steps without significantly impacting the output (maintaining SSIM$>=0.8$). If a further approximation suffices, reducing this quality threshold to SSIM$\geq0.75$ would require relaxations for only $r\in[10,12]$ steps. With this strategy, we can expand interventions strategically while maintaining an acceptable output quality.

\subsection{Using insights for shortcut inference}
We put together our insights for time-optimized inference within diffusion models, aiming to expedite sample generation, critical, for example, in prompt engineering.
\begin{figure}[!b]
    \centering
    \includegraphics[width=0.99\linewidth]{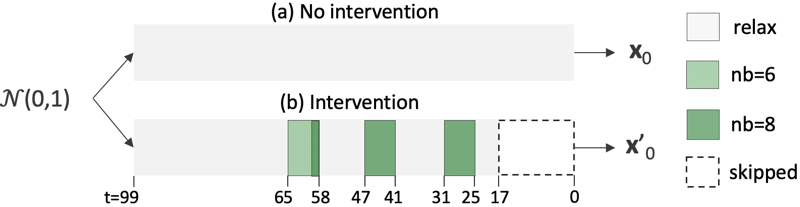}
    \caption{Intervention strategy (b) in comparison to the regular inference workflow (a). }
    \label{fig:strategy}
\end{figure}
In Figure~\ref{fig:strategy}, we apply our insights as follows: starting at $t_{start}=65$, we systematically remove $nb=6$ blocks over $n=5$ steps. Observing that removing $n=7$ steps has minimal impact, we prune $nb=8$ deconvolutional blocks over $n=2$ steps until $t=58$. Building on our earlier findings, we introduce a relaxation period $r$ for ten steps for $t\in(58,47)$, then remove $nb=8$ blocks over $n=7$ steps for $t\in[47,41]$ and iterating through this relax-cut process another time for $t\in(41,25]$.
Our analysis of generation phases revealed that the final time steps primarily involve denoising. Assuming a sufficient estimate of $\textbf{x}'_0$, $\textbf{x}'_{est_{0}}$ in Phase 3, we stop interventions at this stage. Our experiments showed that around $t=18$, this estimate proved reasonably accurate, leading us to skip the final 17 time steps.

\subsubsection{Approximated results}
\begin{figure}[!t]
    \centering
    \includegraphics[width=0.99\linewidth]{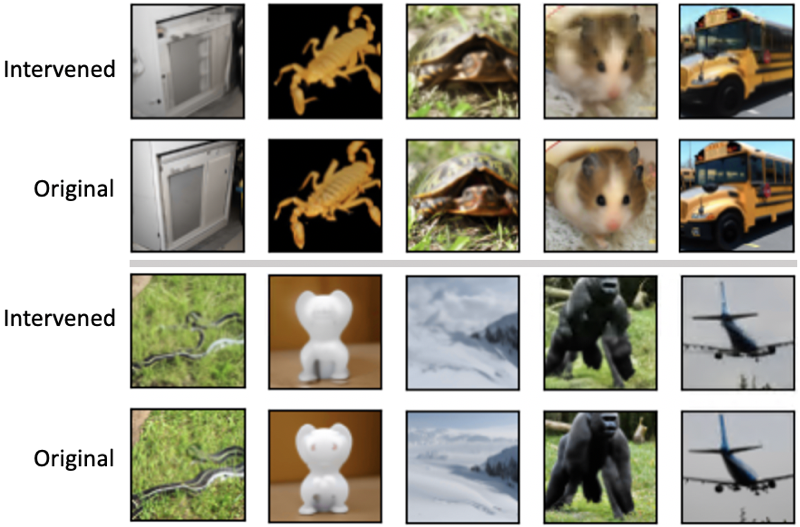}
    \caption{Original non-intervened ($\textbf{x}_0$) and approximated images generated via the intervention strategy described. Approximated images refer to the estimate $\textbf{x}'_{est_{0}}$ at $t=17$ after interventions in Figure~\ref{fig:strategy}. Ten randomly selected image pairs from the test set are shown.}
    \label{fig:approximateimgs}
\end{figure}

\begin{figure}[!b]
    \centering
    \includegraphics[width=0.7\linewidth]{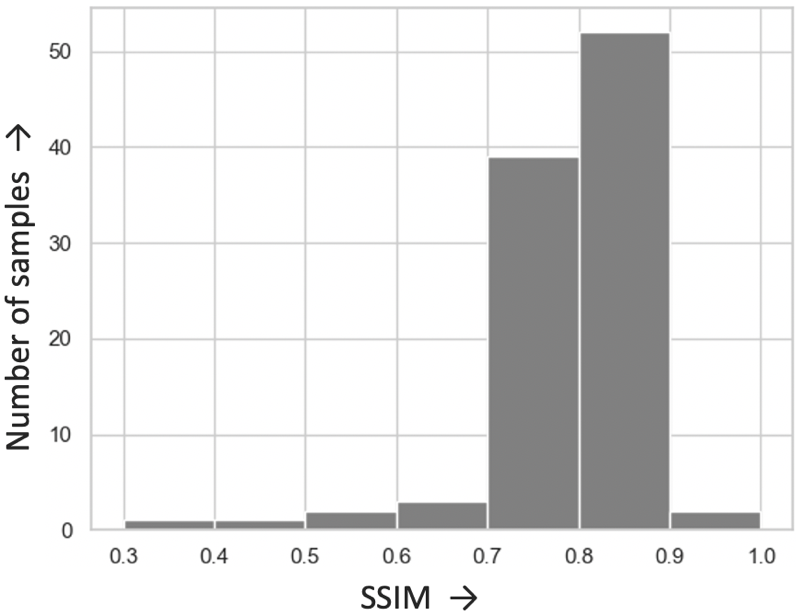}
    \caption{Histogram of SSIM values between the original $\textbf{x}_0$ and the intervened images $\textbf{x}'_{est_{0}}$ at $t=17$ for all 100 samples ($mean=0.795, mad=0.044$). $mad$: Mean absolute deviation. }
    \label{fig:cutstratssimdist}
\end{figure}

: To show how our interventions affect image generation in DDPMs, we run the inference strategy defined in Figure~\ref{fig:strategy} for all 100 samples. As shown in Figure~\ref{fig:approximateimgs}, the intervened samples are quite close in terms of global structure to the original non-intervened outputs. This similarity is also reflected in the distribution of acceptable SSIM values $\in [0.7,0.9]$ across most 100 samples (see Figure~\ref{fig:cutstratssimdist}). The clipping strategy led to a mean percentage decrease of $\sim27\%$ in inference time across the 100 samples. About $5\%$ from the unet interventions and $22\%$ from skipping the last $17$ time steps. 

\begin{figure}
    \centering
    \includegraphics[width=0.95\linewidth]{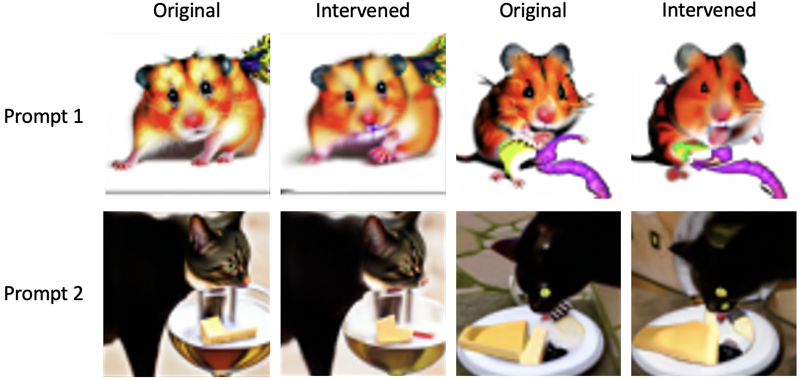}
    \caption{Original non-intervened and approximated images generated via the intervention strategy described previously. Prompt 1: ``A realistic photo of a hamster with a dragon tail''. Prompt 2: ``A sophisticated cat drinking wine with gouda cheese''. Two samples per prompt are shown.}
    \label{fig:prompts_ddpm}
\end{figure}

We further investigate the effectiveness of inference strategy on more complex prompts (see Figure~\ref{fig:prompts_ddpm}).
While there's a trade-off in quality, these approximations offer a global idea of the outputs, enabling users to efficiently iterate through prompts and quickly generate desired results.

While Figure~\ref{fig:strategy} shows one potential shortcut inference method, it is not the sole strategy. Our study aims to shed light on the generation process and elucidate the roles and redundancy patterns of time steps (generative phases), Unet connections, and the model's self-correction capabilities. These insights support further research toward informed optimizations in inference and model design.

\subsection{Conclusions \& Future work}

Our study introduced an analytical method to systematically understand the temporal dynamics and identify potential shortcuts within the inference process of DDPMs. The method eliminated diffusion components to study their influence on the output changes. The systematic interventions on time steps highlighted clear patterns, showcasing significant semantic shifts in initial phases, succeeded by minor semantic changes with coarse denoising, and ending in pure denoising, delineating clear generative stages. Further investigation into Unet connections highlighted their varying utility. The lower layers primarily contributed to semantic changes, while the higher layers were focused on denoising, especially after the initial phase of the generation process. This trend in the utility of connections helped identify redundancies in specific layers, notably toward the end of the first generative phase. 

We highlighted the potential of our insights for accelerated sample generation. Applying these insights to complex prompts demonstrated that while there is some quality trade-off, they sufficiently capture the essence of the model's outputs and could be valuable to efficiently iterate over prompts for desired outputs. Further optimizing the pruned architecture for interference could lead to more accurate results. 

Looking ahead, understanding the generation phases and the purpose and utility trends of Unet connections could guide optimized inference strategies, architectural fine-tuning, and inspire novel models leveraging different phases for improved outcomes. The strategic inference approach could support prompt engineering applications, offering faster and computationally less complex yet sufficiently accurate sample generation.
\bibliography{refs}

\begin{thebibliography}{20}
\providecommand{\natexlab}[1]{#1}

\bibitem[{Baranchuk et~al.(2022)Baranchuk, Rubachev, Voynov, Khrulkov, and
  Babenko}]{segmentation}
Baranchuk, D.; Rubachev, I.; Voynov, A.; Khrulkov, V.; and Babenko, A. 2022.
\newblock Label-Efficient Semantic Segmentation with Diffusion Models.
\newblock In \emph{International Conference on Learning Representations}.

\bibitem[{Bau et~al.(2018)Bau, Zhu, Strobelt, Zhou, Tenenbaum, Freeman, and
  Torralba}]{bau2018gan}
Bau, D.; Zhu, J.-Y.; Strobelt, H.; Zhou, B.; Tenenbaum, J.~B.; Freeman, W.~T.;
  and Torralba, A. 2018.
\newblock Gan dissection: Visualizing and understanding generative adversarial
  networks.
\newblock In \emph{International Conference on Learning Representations}.

\bibitem[{Deja et~al.(2022)Deja, Kuzina, Trzcinski, and
  Tomczak}]{deja2022analyzing}
Deja, K.; Kuzina, A.; Trzcinski, T.; and Tomczak, J. 2022.
\newblock On analyzing generative and denoising capabilities of diffusion-based
  deep generative models.
\newblock \emph{Advances in Neural Information Processing Systems}, 35:
  26218--26229.

\bibitem[{Ho, Jain, and Abbeel(2020)}]{ho2020denoising}
Ho, J.; Jain, A.; and Abbeel, P. 2020.
\newblock Denoising diffusion probabilistic models.
\newblock \emph{Advances in neural information processing systems}, 33:
  6840--6851.

\bibitem[{Jolicoeur-Martineau et~al.(2022)Jolicoeur-Martineau, Li,
  Pich{\'e}-Taillefer, Kachman, and Mitliagkas}]{jolicoeur-martineau2022gotta}
Jolicoeur-Martineau, A.; Li, K.; Pich{\'e}-Taillefer, R.; Kachman, T.; and
  Mitliagkas, I. 2022.
\newblock Gotta Go Fast When Generating Data with Score-Based Models.

\bibitem[{Karras et~al.(2022)Karras, Aittala, Aila, and
  Laine}]{karras2022elucidating}
Karras, T.; Aittala, M.; Aila, T.; and Laine, S. 2022.
\newblock Elucidating the design space of diffusion-based generative models.
\newblock \emph{Advances in Neural Information Processing Systems}, 35:
  26565--26577.

\bibitem[{Kawar et~al.(2022)Kawar, Elad, Ermon, and Song}]{inverseprob}
Kawar, B.; Elad, M.; Ermon, S.; and Song, J. 2022.
\newblock Denoising diffusion restoration models.
\newblock \emph{Advances in Neural Information Processing Systems}, 35:
  23593--23606.

\bibitem[{Kong and Ping(2021)}]{kong2021fast}
Kong, Z.; and Ping, W. 2021.
\newblock On Fast Sampling of Diffusion Probabilistic Models.
\newblock In \emph{ICML Workshop on Invertible Neural Networks, Normalizing
  Flows, and Explicit Likelihood Models}.

\bibitem[{Nichol et~al.(2022)Nichol, Dhariwal, Ramesh, Shyam, Mishkin, McGrew,
  Sutskever, and Chen}]{nichol2021glide}
Nichol, A.; Dhariwal, P.; Ramesh, A.; Shyam, P.; Mishkin, P.; McGrew, B.;
  Sutskever, I.; and Chen, M. 2022.
\newblock {GLIDE}: Towards Photorealistic Image Generation and Editing with
  Text-Guided Diffusion Models.
\newblock In \emph{Proceedings of the 39th International Conference on Machine
  Learning}, 16784--16804.

\bibitem[{Nichol and Dhariwal(2021)}]{nichol2021improved}
Nichol, A.~Q.; and Dhariwal, P. 2021.
\newblock Improved denoising diffusion probabilistic models.
\newblock In \emph{International Conference on Machine Learning}, 8162--8171.
  PMLR.

\bibitem[{Rombach et~al.(2022)Rombach, Blattmann, Lorenz, Esser, and
  Ommer}]{stablediff}
Rombach, R.; Blattmann, A.; Lorenz, D.; Esser, P.; and Ommer, B. 2022.
\newblock High-resolution image synthesis with latent diffusion models.
\newblock In \emph{Proceedings of the IEEE/CVF conference on computer vision
  and pattern recognition}, 10684--10695.

\bibitem[{Ronneberger, Fischer, and Brox(2015)}]{ronneberger2015u}
Ronneberger, O.; Fischer, P.; and Brox, T. 2015.
\newblock U-net: Convolutional networks for biomedical image segmentation.
\newblock In \emph{Medical Image Computing and Computer-Assisted
  Intervention--MICCAI 2015: 18th International Conference, Munich, Germany,
  October 5-9, 2015, Proceedings, Part III 18}, 234--241. Springer.

\bibitem[{Shih et~al.(2023)Shih, Belkhale, Ermon, Sadigh, and
  Anari}]{shih2023parallel}
Shih, A.; Belkhale, S.; Ermon, S.; Sadigh, D.; and Anari, N. 2023.
\newblock Parallel Sampling of Diffusion Models.
\newblock \emph{arXiv preprint arXiv:2305.16317}.

\bibitem[{Sohl-Dickstein et~al.(2015)Sohl-Dickstein, Weiss, Maheswaranathan,
  and Ganguli}]{sohl2015deep}
Sohl-Dickstein, J.; Weiss, E.; Maheswaranathan, N.; and Ganguli, S. 2015.
\newblock Deep unsupervised learning using nonequilibrium thermodynamics.
\newblock In \emph{International conference on machine learning}, 2256--2265.
  PMLR.

\bibitem[{Song, Meng, and Ermon(2021)}]{ddim}
Song, J.; Meng, C.; and Ermon, S. 2021.
\newblock Denoising Diffusion Implicit Models.
\newblock In \emph{International Conference on Learning Representations}.

\bibitem[{Wyatt et~al.(2022)Wyatt, Leach, Schmon, and
  Willcocks}]{wyatt2022anoddpm}
Wyatt, J.; Leach, A.; Schmon, S.~M.; and Willcocks, C.~G. 2022.
\newblock Anoddpm: Anomaly detection with denoising diffusion probabilistic
  models using simplex noise.
\newblock In \emph{Proceedings of the IEEE/CVF Conference on Computer Vision
  and Pattern Recognition}, 650--656.

\bibitem[{Yang et~al.(2022)Yang, Zhang, Song, Hong, Xu, Zhao, Zhang, Cui, and
  Yang}]{yang2022diffusionsurvey}
Yang, L.; Zhang, Z.; Song, Y.; Hong, S.; Xu, R.; Zhao, Y.; Zhang, W.; Cui, B.;
  and Yang, M.-H. 2022.
\newblock Diffusion models: A comprehensive survey of methods and applications.
\newblock \emph{ACM Computing Surveys}.

\bibitem[{Zhang and Chen(2023)}]{deis}
Zhang, Q.; and Chen, Y. 2023.
\newblock Fast Sampling of Diffusion Models with Exponential Integrator.
\newblock In \emph{The Eleventh International Conference on Learning
  Representations}.

\bibitem[{Zhang, Tao, and Chen(2023)}]{zhang2023gddim}
Zhang, Q.; Tao, M.; and Chen, Y. 2023.
\newblock g{DDIM}: Generalized denoising diffusion implicit models.
\newblock In \emph{The Eleventh International Conference on Learning
  Representations}.

\bibitem[{Zheng et~al.(2023)Zheng, Nie, Vahdat, Azizzadenesheli, and
  Anandkumar}]{zheng2023fast}
Zheng, H.; Nie, W.; Vahdat, A.; Azizzadenesheli, K.; and Anandkumar, A. 2023.
\newblock Fast sampling of diffusion models via operator learning.
\newblock In \emph{International Conference on Machine Learning}, 42390--42402.
  PMLR.

\end{thebibliography}
\newpage
\clearpage
\appendix
\section{Appendix}

\begin{figure}[!h]
    \centering
    \includegraphics[width=0.8\linewidth]{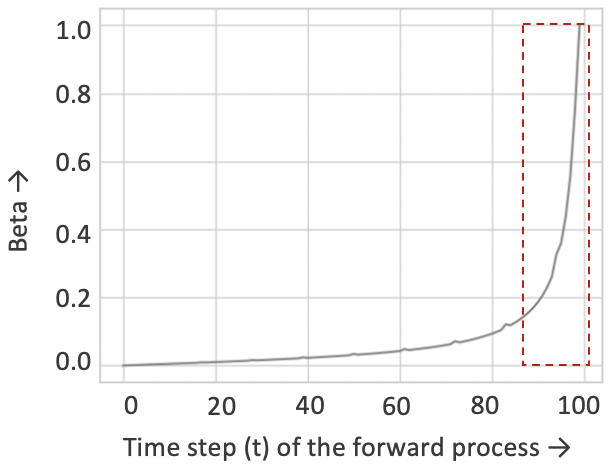}
    \caption{Beta schedule used for the base model in GLIDE~\cite{nichol2021glide}. Note that the schedule above refers to a schedule for spaced diffusion with 100 time steps. Notably, the highlighted red box signals a significant noise pick up in noise beyond $t>85$, aligning with substantial semantic shifts observed during these time steps in our experiments.}
    \label{fig:beta_scheduler}
\end{figure}

\begin{figure}[!h]
    \centering
    \includegraphics[width=0.8\linewidth]{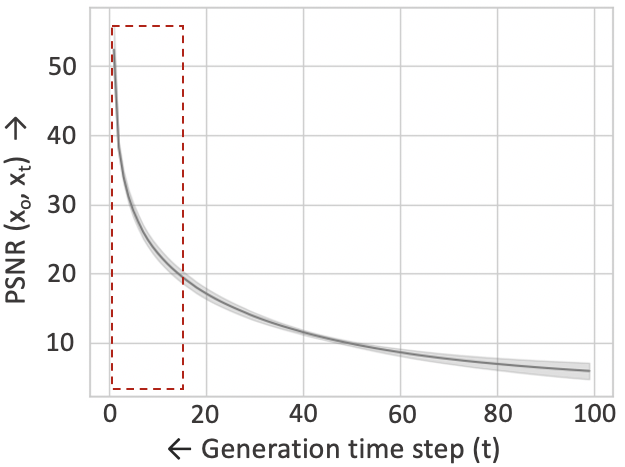}
    \caption{
    Analyzing time step redundancy via PSNR between intermediate $\textbf{x}_t$ and final $\textbf{x}_0$: The mean $\pm$ 2 standard deviations across 100 instances is shown. Notably marked within the red box, the area around time steps $t\in[15,18]$ shows a sharp rise in PSNR values, indicating rapid convergence of $\textbf{x}_t$ towards $\textbf{x}_0$. This observation strongly suggests the lower significance of these final time steps in the overall process, potentially indicating their redundancy.}
    \label{fig:xt_estimate}
\end{figure}
\begin{figure}[!b]
    \centering
    \includegraphics[width=0.8\linewidth]{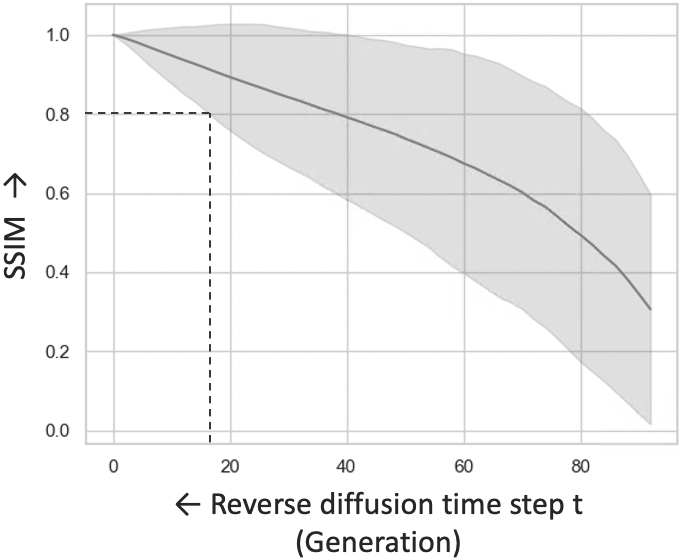}
    \caption{
    Assessing redundancy of time steps via SSIM analysis of $\textbf{x}_{est_{0}}$: The mean $\pm$ 2 standard deviations across 100 instances are shown. In phase 3, primarily focused on denoising, our question is whether $\textbf{x}_0$ estimates at $t\le25$ adequately preserve global image structure. Notably, analyzing SSIM comparisons between $\textbf{x}_0$ and the estimated $\textbf{x}_0$ at each time step $\textbf{x}_{est_0}^t$ shows that $\textbf{x}_{est_0}^{t=18}$ offers a accurate representation. This indicates that subsequent time steps ($t\le18$) might not be essential for achieving a satisfactory $\textbf{x}_0$ approximation.}
    \label{fig:x0_est_goodness}
\end{figure}

\begin{figure}[!h]
    \centering
    \includegraphics[width=0.8\linewidth]{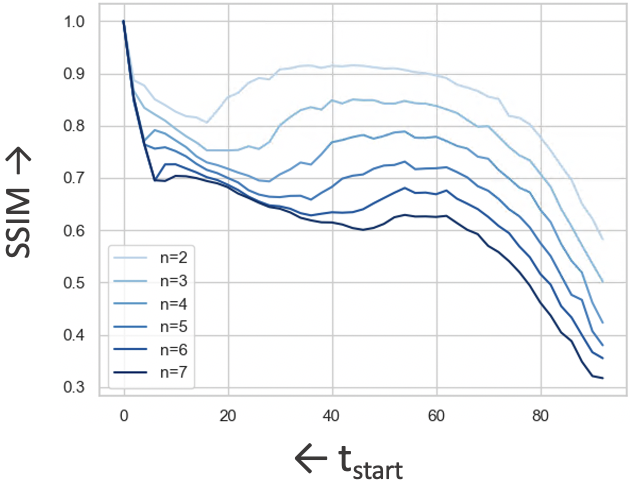}
    \caption{The impact of removed time steps $n$ on SSIM. The plot shows the SSIM between the original $\textbf{x}_0$ and the intervened outputs $\textbf{w}'_0$ after removing $n$ time steps, starting at step $t_{start}$ on the x-axis. The mean $\pm$ SSIM across 100 instances is shown. This exploration aims to identify perceivable changes. Notably, for an SSIM threshold of 0.8, at $n=5$, most intervened outputs exhibit SSIM values below the threshold, indicating noticeable visual alterations for exploration time step purposes.}
    \label{fig:nvarytimesteps}
\end{figure}

\begin{figure*}[!h]
    \centering
    \begin{subfigure}[t]{0.5\textwidth}
        \centering
        \includegraphics[width=0.95\linewidth]{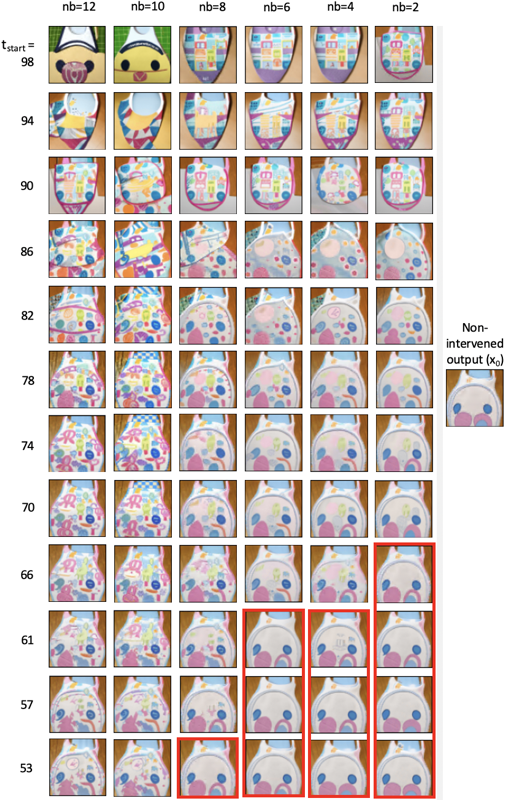}
        \caption{$t_{start}\in[50,100]$}
    \end{subfigure}%
    \begin{subfigure}[t]{0.5\textwidth}
        \centering
        \includegraphics[width=0.95\linewidth]{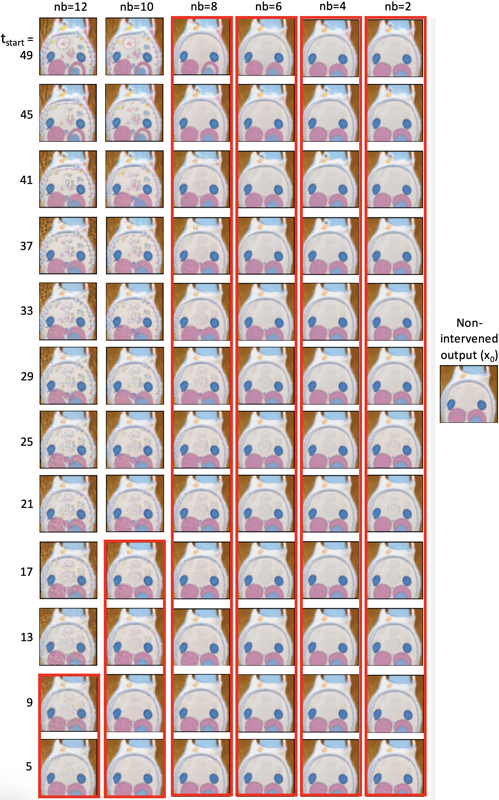}
        \caption{$t_{start}\in[0,49]$}
    \end{subfigure}
    \caption{
    The impact of deconvolutional block removal on the intervened $\textbf{b}'_0$: The figure shows $\textbf{b}'_0$ obtained by removing $nb$ blocks across $n=5$ consecutive time steps from $t{start}$ to $t_{start}-5$. In red are intervened images exhibiting visual similarity to the non-intervened output ($\textbf{x}_0$). An interesting pattern emerges from the highlighted boxes, highlighting a sequential redundancy of blocks, noticeable in a bottom-to-top trend. }
\end{figure*}
\begin{figure}[!h]
    \centering
    \includegraphics[width=0.99\linewidth]{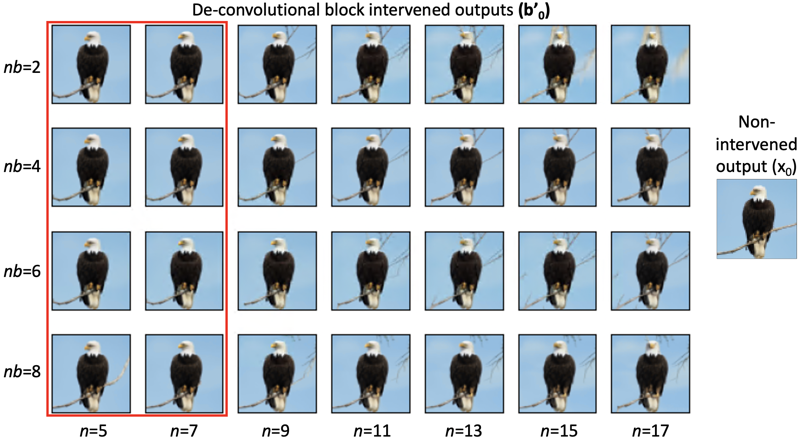}
    \caption{The effects of the number of time steps intervened on $\textbf{b}'_0$. Above is the impact of removing $nb$ blocks across consecutive steps ($n$) on $\textbf{b}'_0$. Minimal changes persist up to $n=7$ (see red box), beyond which clear semantic differences begin to emerge.}
    \label{fig:varywindow_images}
\end{figure}

\begin{figure}[!h]
    \centering
    \includegraphics[width=0.46\linewidth]{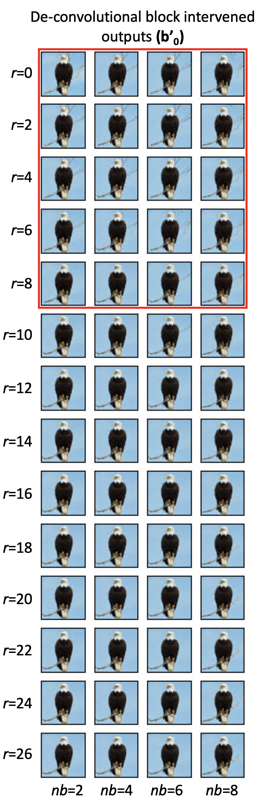}
    \caption{Relaxation steps: The observed deconvolution block after removing $nb$ blocks across $n=7$ consecutive time steps, followed by a relaxation phase of $r$ steps, and another round of removing $nb$ blocks for $n=7$ time steps. Limited semantic changes are evident after a relaxation period exceeding $r=10$ steps (highlighted in the red box).}
    \label{fig:cut-relax-cut-img}
\end{figure}

\begin{figure}[!t]
    \centering
    \includegraphics[width=0.99\linewidth]{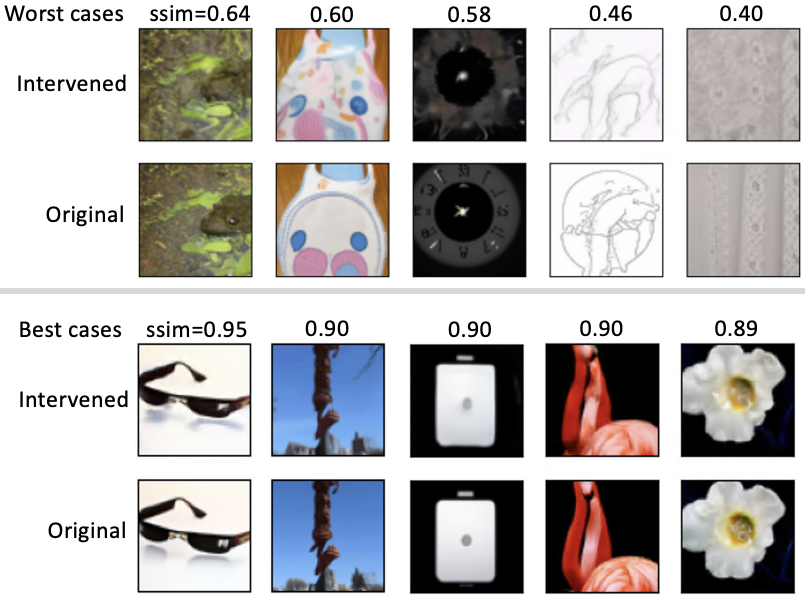}
    \caption{Comparison between the original non-intervened ($\textbf{x}_0$) and approximated images ($\textbf{x}'_0$) generated using the described intervention strategy. The figure displays the worst and best images from the 100-sample test set. Notably, the images categorized as worst cases represent the tail end of the distribution seen in Figure 13. Specifically, the three images in the far right of the worst-case fail primarily due to the presence of fine lines. This issue arises due to our clipping strategy's emphasis on global approximations rather than preserving finer details. }
    \label{fig:bestworst_clippedsamples}
\end{figure}
\end{document}